\documentclass[journal,twocolumn]{IEEEtran}

\usepackage{amsmath}
\usepackage{subfigure}
\usepackage{caption}
\usepackage{algorithm}
\usepackage{algpseudocode}
\usepackage{setspace}
\usepackage{color,hyperref}

\usepackage{graphicx}
\usepackage{amsmath,amssymb} 
\usepackage{epsfig}
\usepackage{amsmath}
\usepackage{amssymb}
\usepackage{colortbl}
\usepackage{soul}
\usepackage{multirow}
\usepackage{pifont}
\usepackage{bm}

\usepackage[table ]{ xcolor}
\soulregister\cite7
\soulregister\ref7

\newcommand{\xmark}{\ding{55}}
\newcommand{\rmark}{\ding{52}}

\definecolor{mygray}{gray}{.9}

\newcolumntype{I}{!{\vrule width 1.2pt}}
\newlength\savedwidth
\newcommand\whline{\noalign{\global\savedwidth\arrayrulewidth
		\global\arrayrulewidth 1.25pt}%
	\hline
	\noalign{\global\arrayrulewidth\savedwidth}}

\ifCLASSINFOpdf
\else
\fi

\definecolor{darkblue}{rgb}{0.0,0.0,1.0}
\hypersetup{colorlinks,breaklinks,
	linkcolor=darkblue,urlcolor=darkblue,
	anchorcolor=darkblue,citecolor=darkblue}

\begin{document}
\setul{}{1.5pt}

\title{Domain-adaptive Crowd Counting via High-quality Image Translation and Density Reconstruction }

\author{
        Junyu~Gao,~\IEEEmembership{Member,~IEEE,}
        Tao~Han,~\IEEEmembership{~Student Member,~IEEE,}
        ~Yuan~Yuan,~\IEEEmembership{~Senior Member,~IEEE,}
        and Qi~Wang,~\IEEEmembership{~Senior Member,~IEEE}
	\thanks{
	J. Gao, T. Han, Y. Yuan and Q. Wang are with the School of Artificial Intelligence, Optics and Electronics (iOPEN), Northwestern Polytechnical University, Xi'an 710072, Shaanxi, China. E-mails: gjy3035@gmail.com, hantao10200@mail.nwpu.edu.cn, crabwq@gmail.com, y.yuan1.ieee@gmail.com. Q. Wang is the corresponding author.

	\copyright 20XX IEEE. Personal use of this material is permitted. Permission from IEEE must be obtained for all other uses, in any current or future media, including reprinting/republishing this material for advertising or promotional purposes, creating new collective works, for resale or redistribution to servers or lists, or reuse of any copyrighted component of this work in other works.}
}
\markboth{{IEEE} Transactions on Neural Networks and Learning Systems}%
{Shell \MakeLowercase{\textit{et al.}}: Bare Demo of IEEEtran.cls for Journals}
\maketitle

\begin{abstract}

Recently, crowd counting using supervised learning achieves a remarkable improvement. Nevertheless, most counters rely on a large amount of manually labeled data. With the release of synthetic crowd data, a potential alternative is transferring knowledge from them to real data without any manual label. However, there is no method to effectively suppress domain gaps and output elaborate density maps during the transferring. To remedy the above problems, this paper proposes a Domain-Adaptive Crowd Counting (DACC) framework, which consists of a high-quality image translation and density map reconstruction. To be specific, the former focuses on translating synthetic data to realistic images, which prompt the translation quality by segregating domain-shared/independent features and designing content-aware consistency loss. The latter aims at generating pseudo labels on real scenes to improve the prediction quality. Next, we retrain a final counter using these pseudo labels. Adaptation experiments on six real-world datasets demonstrate that the proposed method outperforms the state-of-the-art methods.
	
\end{abstract}


\section{Introduction}
\setulcolor{red}
\label{intro}
Crowd counting is usually treated as a pixel-level estimation problem, which predicts the density value for each pixel and sums the entire prediction map as a final counting result. A pixel-wise density map produces more detailed information than a single number for a complex crowd scene. In addition, it also boosts other highly semantic crowd analysis (group detection \cite{wang2018detecting,li2017multiview,li2017locality}, crowd segmentation \cite{kang2014fully}, public management \cite{zhou2013measuring}, \emph{etc.}) or video surveillance tasks (video summarization \cite{muhammad2018efficient,zhao2018hsa,DBLP:journals/tnn/ZhaoLL20} and abnormal detection \cite{yuan2018structured}). Recently, benefiting from the powerful capacity of deep learning, there is a significant promotion in the field of counting. However, currently released datasets are too small to satisfy the mainstream deep-learning-based methods \cite{zhang2015cross,sam2017switching,babu2018divide,liu2018decidenet,li2018csrnet,shi2019revisiting}. The main reason is that constructing a large-scale crowd counting dataset is extremely demanding, which needs many human resources \cite{gao2020nwpu}.

To handle the scarce data problem, many researchers pay attentions to data generation. Exploiting computer graphics to render photo-realistic crowd scenes becomes an alternative to generate a large-scale dataset \cite{wang2019learning}. Unfortunately, due to the differences between the synthetic and real worlds (also named as ``domain shifts/gap''), there is an obvious performance degradation when applying the synthetic crowd model to the real world.
For reducing the domain shifts, Wang \emph{et al.} \cite{wang2019learning} are the first to propose a crowd counting via domain adaptation method based on CycleGAN \cite{zhu2017unpaired}, which translates synthetic data to photo-realistic scenes and then apply the trained model in the wild. In this paper, we also focus on Domain-Adaptive Crowd Counting (DACC), which attempts to transfer the useful knowledge for crowd counting from a source domain (the synthetic data) to a target domain (the real world).

\begin{figure*}[t]
	\centering
	\includegraphics[width=\textwidth]{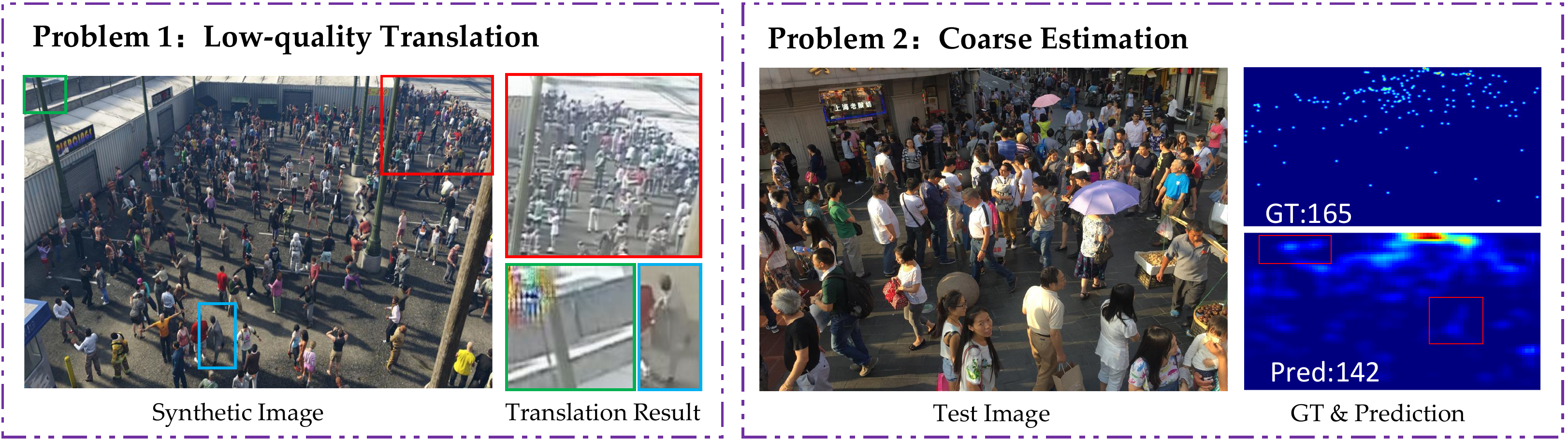}
	\caption{Existing problems in the current domain-adaptive crowd counting.  }\label{Fig-intro}
\end{figure*}

However, there are \textbf{two problems} in the CycleGAN-style \cite{zhu2017unpaired,hoffman2017cycada,wang2019learning,chen2019crdoco} adaptation methods: 1) output some distorted translations and lose many textures and local structured patterns (these features are key characteristics for congested crowd scenes), which produce coarse density map. The left box in Fig. \ref{Fig-intro} shows the three types of false cases (Red: lost textures, Green: distorted data, and Blue: lost local pattern). 2) mistakenly estimate response values for unseen background objects in the target domain so that the prediction map is very coarse and inaccurate. The right box in Fig. \ref{Fig-intro} demonstrates some mis-estimations of the background.

For the first problem, the main reason is that CycleGAN only classifies the translated and recalled results at the image level and treats image translation as an entire process. In practice, we find that different domains have common crowd contents, namely person's structure features and crowd distribution patterns, which is regarded as ``domain-shared features''. Besides, different domains have own unique scene attributes, named as ``domain-independent features'', which may be caused by different factors such as backgrounds, sensors' setting. Motivated by this discovery, we propose a two-step chain architecture to segregate the two types of features, named as Inter-domain Features Segregation (IFS). It firstly extracts domain-shared features $f$. Next, by decorating $f$ with the domain-independent features of domain $\mathcal{T}$, IFS reconstructs the like-$\mathcal{T}$ images. For further maintaining the local patterns and texture features, we carefully design multi-scale adversarial translation loss and content-aware loss. Compared with the traditional adversarial loss and Cycle loss, the proposed losses can significantly  reduce distortions and retain image contents during the translation.

For the second problem, we present a re-training scheme base on the density reconstruction. In the counting field, the ground-truth of density map is generated by using a Gaussian kernel from the head position. According to this prior, we attempt to find the most likely locations of heads by comparing the similarity between the coarse map and the standard Gaussian kernel. Consequently, pseudo density maps are reconstructed. Then a final counter is trained on the target images and the pseudo maps, which performs better in the real world than the coarse model.

As a summary, the key contributions of this paper are:

\begin{enumerate}
	\item[1)] Propose a two-step image translation to segregate inter-domain features, and design two effective types of losses, which can extract/retain crowd contents and yield high-quality photo-realistic crowd images.
	\item[2)] Exploit Gaussian prior to reconstruct pseudo labels according to the coarse results. Based on them, retrain a fine counter to further enhance the density quality and counting performance.
	\item[3)] The proposed method outperforms the state-of-the-art results in the domain-adaptive crowd counting from synthetic data to the real world. 	
\end{enumerate}

\section{Related Works}


\subsection{Crowd Counting}

\textbf{Supervised Learning.} Early methods for crowd counting focus on extracting hand-crafted features (such as Harr \cite{viola2004robust}, HOG \cite{dalal2005histograms}, texture features \cite{brostow2006unsupervised}, etc.) to regress the number of people \cite{ryan2009crowd,lempitsky2010learning,idrees2013multi}. Recently, many object counting researches are based on CNN methods. Some researchers design network structures to enhance multi-scale feature extraction capabilities \cite{zhang2016single,onoro2016towards}.  Zhang \emph{et al.} \cite{zhang2016single} propose a multi-column CNN by combining different kernel sizes. Onoro-Rubio and  L{\'o}pez-Sastre \cite{onoro2016towards} present a multi-scale Hydra CNN, which performs the density prediction in different scenes. Some works \cite{sindagi2017generating,liu2019context,gao2019scar,gao2020pcc} exploit contextual information to boost counting performance. Sindagi \emph{et al.} \cite{sindagi2017generating} extract global and local feature to aid the density estimation, and Liu \emph{et al.} \cite{liu2019context} present a context-aware CNN, designing a multi-stream with different respective fields after a VGG backbone.  The rest works \cite{idrees2018composition,jiang2019crowd,liu2019crowd} fuse multi-stage features to achieve accurate counting. Idrees \emph{et al.} \cite{idrees2018composition} combine the results of different stages to predict the density map and head localization. Jiang \emph {et al.} \cite{jiang2019crowd} design a trellis encoder-decoder architecture to incorporate the features from multiple decoding paths. Liu \emph{et al.} \cite{liu2019crowd} present a Structured Feature Enhancement Module (SFEM) using Conditional Random Field (CRF) to refine the features of different stages.

\textbf{Counting for Scarce Data.} In addition to the aforementioned supervised methods, some approaches dedicate to handling the problem of scarce data. Wang \emph{et al.} \cite{wang2019learning} construct a large-scale synthetic crowd dataset, including more than $15,000$ images, $\sim\!7.5$ million instances. Recently, two real-world crowd datasets are released, namely JHU-CROWD \cite{sindagi2019pushing} ($4,250$ images, $\sim\!1.1$ million instances) and Crowd Surveillance \cite{yan2019perspective} ($13,945$ images, $\sim\!0.4$ million annotations). By comparing them, the amount of labeled real data is far from that of synthetic data. Besides, collecting and annotating real data is an expensive and difficult assignment. Thus, some researchers remedy this problem from the methodology. Liu \emph{et al.} \cite{liu2018leveraging} propose a self-supervised ranking scheme as an auxiliary to improve performance. Sam \emph{et al.} \cite{sam2019almost} present an almost unsupervised method, of which 99.9\% parameters in the proposed auto-encoder is trained without any label. Olmschenk \emph{et al.} \cite{olmschenk2019generalizing} enlarge the data by utilizing Generative Adversarial Networks (GAN). To fully escape from manually labeled data and simultaneously attain an accepted result, Wang \emph{et al.} \cite{wang2019learning} present a crowd counting via domain adaptation method, which is easy to land in practice from the perspectives of performance and costs.

\subsection{Domain-adaptive Vision Tasks}

Considering that there are not many works about domain adaptation in crowd counting, thus this section reviews other applications, such as classification, segmentation, etc. Some methods \cite{long2015learning,bousmalis2016domain,DBLP:conf/icml/LongZ0J17} adopt the Maximum Mean Discrepancy \cite{gretton2012kernel} to alleviate domain shift in the field of image classification. After some synthetic segmentation datasets \cite{richter2016playing,RosCVPR16} are released, a few works \cite{hoffman2016fcns,sankaranarayanan2018learning,han2020focus,gao2019feature} adopt adversarial learning to reduce the pixel-wise domain gap. Benefiting from the power of CycleGAN \cite{zhu2017unpaired}, some scholars \cite{hoffman2017cycada,chen2019crdoco} utilize it to translate synthetic images to realistic data. Recently, some researchers attempt to disentangle the image content and style to translate images \cite{gonzalez2018image,chang2019all,lu2019unsupervised,li2019cross}. From these works for image classification and segmentation tasks, they only focus on global styles and object-level structures. For counting task, in addition to the style and structures, we also devote to preserving the consistency of local pattern and textures. Thus, we start from the translation architecture and loss designation to achieve our goal.

\textbf{Different from the traditional segregation methods} \cite{8578260,chang2019all} that extract a simple domain code to generate images: in order to reconstruct the image details, we uses two different generators that contains a large number of neurons. The richer domain-independent attributes are stored in the generators than the simple domain code.

\textbf{Different from the previous Cycle-Consistent methods} \cite{zhu2017unpaired,hoffman2017cycada,chen2019crdoco} that only constrain the original image and the recalled image by a cycle-consistent loss: to maintain the key content during the translation process, we should regularize the original image and the translated image. To this end, we propose a content-aware consistency loss to guarantee translated data do not lose the original image content, local pattern and texture information.

\textbf{Different from the previous feature-level adversarial learning algorithms} \cite{gao2019feature,han2020focus} that directly learn the domain-invariant features: the proposed method is based on high-quality image translation, which is more interpretable. Besides, it can be treated as data augmentation. Cai \emph{et al.} \cite{cai2021leveraging} propose a two-stage domain adaptation method, which first utilizes adversarial learning in multi-level feature to strengthen target domain's adaptability, and then uses the predicted density maps in the first stage as the pseudo-labels to retrain the counter.

\section{Our Method}
\label{method}

\begin{figure*}[htbp]
	\centering
	\includegraphics[width=\textwidth]{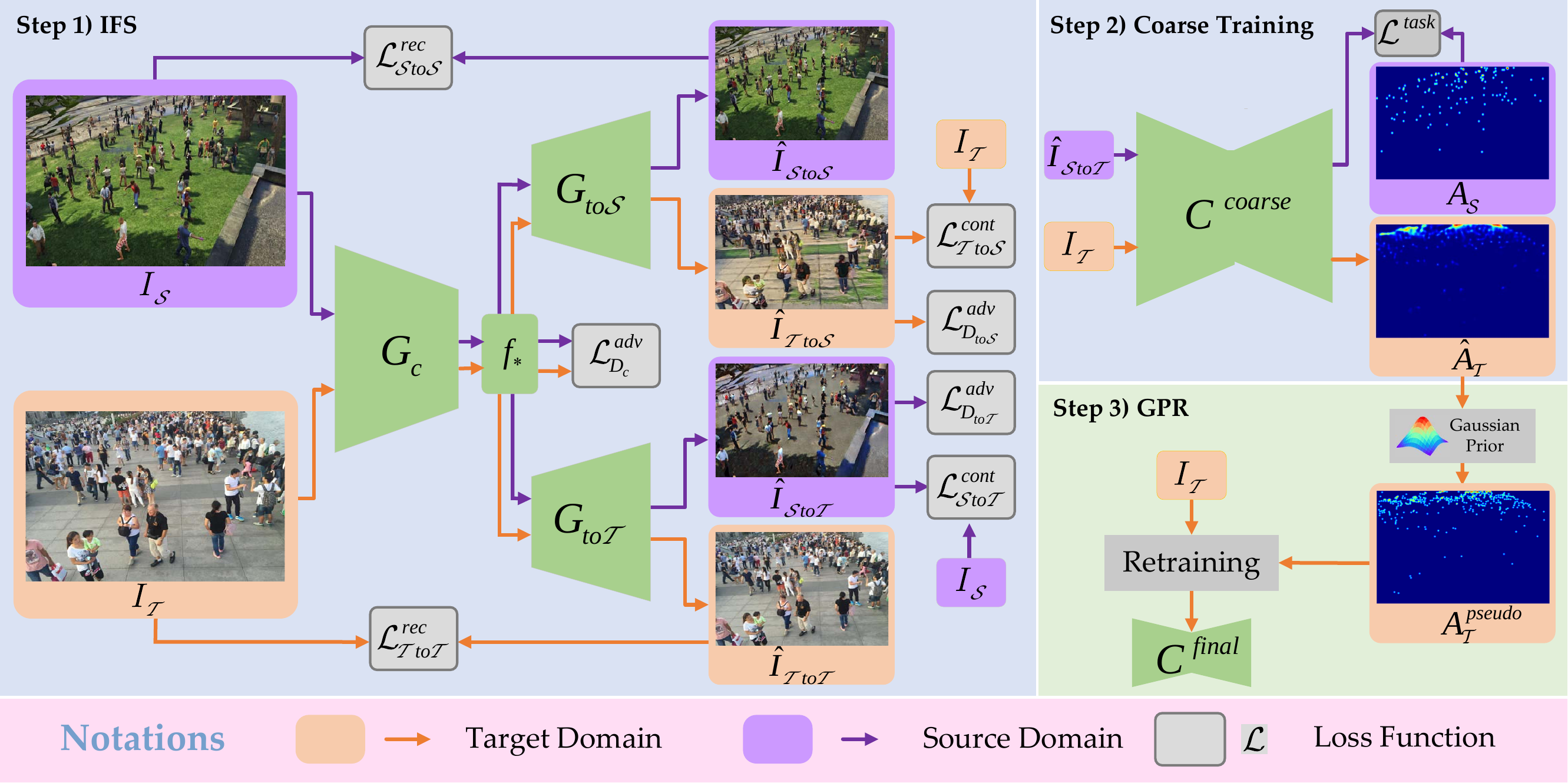}
	\caption{The flowchart of our proposed method, which consists of three components: 1) IFS translates ${I_\mathcal{S}}$ to ${I_{\mathcal{S}to\mathcal{T}}}$; 2) Train the coarse counter $\bm{C}^{coarse}$ using ${I_{\mathcal{S}to\mathcal{T}}}$ and $A_\mathcal{S}$; 3) After $\bm{C}^{coarse}$ converges via iteratively optimizing Step 1) and 2), reconstruct the pseudo map $A_\mathcal{T}^{pseudo}$ from $\bm{C}^{coarse}$'s predictions $\hat A_\mathcal{T}$ and retrain the final counter $\bm{C}^{final}$ using ${I_{\mathcal{T}}}$ and $A_\mathcal{T}^{pseudo}$. Limited by the paper space, the three discriminators are not shown in the figure.  } \label{Fig-overview}
\end{figure*}

Here, the proposed DACC is explained from the perspective of data flow. Specifically, a source domain provides crowd images ${I_\mathcal{S}}$ with the labeled density maps $A_\mathcal{S}$; and a target domain only provides images ${I_\mathcal{T}}$. The purpose is to get the prediction density maps $\hat A_\mathcal{T}$ according to given the ${I_\mathcal{S}}$, $A_\mathcal{S}$ and ${I_\mathcal{T}}$. To help the reader understand, some of the symbol used behind are concluded in Table \ref{Table-annotation}

\begin{table}[htbp]
	\centering
	\caption{Some beforehand annotations of involved symbol.}
	\setlength{\tabcolsep}{1.6mm}{
		\begin{tabular}{cIc}
            \whline
			Symbol& Explanation \\
			\whline
			${\mathcal{S}}$   &  source domain (synthetic data) \\
			\hline
			${\mathcal{T}}$   &  target  domain (real-world data)\\
			\hline	
			$\bm{G}_{c}$       & domain-shared feature extractor (see Fig. 2)\\
			\hline	
			$\bm{G}_{to\mathcal{S}}$ & source domain decoder (see Fig. 2)\\
			\hline
			$\bm{G}_{to\mathcal{T}}$ & target domain decoder  (see Fig. 2)\\
            \hline
            $\bm{D}_{c}$ &  domain-shared feature discriminator (see Fig. 2)\\
             \hline
            $\bm{D}_{to\mathcal{S}}$ & source domain discriminator (see Fig. 2)\\
             \hline
            $\bm{D}_{to\mathcal{T}}$ &  target domain discriminator  (see Fig. 2)\\
			\whline
		\end{tabular}
	}\label{Table-annotation}
\end{table}

\subsection{High-quality Image Translation for Crowd Counting}
Image translation aims to translate source images ${I_{\mathcal{S}}}$ to like-target data ${\hat I_{\mathcal{S}to\mathcal{T}}}$. At the same time, the latter is supposed to contain the key crowd contents of the former. Inspired by the disentangled representation \cite{gonzalez2018image,chang2019all}, we propose an Inter-domain Features Segregation (IFS) framework to separate the crowd contents and domain-independent attributes. Finally, exploiting the translated images and source labels, we train a coarse crowd counter.

\vspace{0.15cm}
\noindent\textbf{3.1.1 \,\,Inter-domain Features Segregation}
\vspace{0.15cm}

\label{IFS}

\noindent\textbf{Assumption.}\quad For crowd scenes of different domains, some essential contents are shared, such as the structure information of persons, the arrangement of congested crowds. Meanwhile, each domain has its private attributes, such as different backgrounds, image styles, viewpoints. Thus, we assume that \emph{a source domain shares a latent feature space with any other target domain, and each domain has its independent attribute}.

\noindent\textbf{Model Overview.}\quad Based on this assumption, the purpose of IFS is supposed to separate common crowd contents and private attributes without overlapping. It consists of two components, a domain-shared features extractor $\bm{G}_{c}$ and two domain-specific decoders $\bm{G}_{to\mathcal{S}}$ and $\bm{G}_{to\mathcal{T}}$ for source and target domains. To separate two types of features, we design three corresponding adversarial discriminators for them. The discriminators attempt to distinguish which domain the outputs of $\bm{G}_{c}$, $\bm{G}_{to\mathcal{S}}$ and $\bm{G}_{to\mathcal{T}}$ come from. By optimizing generators and discriminators in turns, $\bm{G}_{c}$ can extract domain-shared features, and $\bm{G}_{to\mathcal{S}}$, $\bm{G}_{to\mathcal{T}}$ can reconstruct source domain-like or target domain-like crowd scenes according to the outputs of feature extractor. Consequently, the domain-shared features are extracted explicitly and the domain-specific features are implicitly contained in source domain decoder and target domain decoder.

\noindent\textbf{Domain-shared Features Extractor $\bm{G}_{c}$ .}\quad Based on the above assumption, it is important to ensure that feature extractor extracts similar feature distributions for the samples from different domains (namely $i_\mathcal{S} \in I_\mathcal{S}$ and $i_\mathcal{T} \in I_\mathcal{T}$). To this end, we introduce a feature-level adversarial learning for the $f_{\mathcal{S}}$ and $f_\mathcal{T}$ produced by the features extractor, of which is corresponding to source domain and target domain, respectively. Specifically, training a discriminator $\bm{D}_{c}$ to distinguish whether the features come from source domain or target domain. At the same time, updating the parameters of feature extractor to fool $\bm{D}_{c}$ by using the loss of the inverse discrimination result. Consequently, $f_\mathcal{S}$ and $f_\mathcal{T}$ are very similar and share the same feature space.

\noindent\textbf{Domain-specific Decoders $\bm{G}_{to\mathcal{S}}$ and $\bm{G}_{to\mathcal{T}}$.}\quad The proposed $\bm{G}_{c}$ can extract the features that share the same feature space, but it does not mean that they are key contents mentioned in Assumption. Thus, we propose two domain-specific decoders for domain $\mathcal{S}$ and $\mathcal{T}$, which reconstruct images like own domain according to the outputs of feature extractor. On the one hand, this process encourages feature extractor to extract effective domain-shared features. On the other hand, it makes $\bm{G}_{to\mathcal{S}}$ and $\bm{G}_{to\mathcal{T}}$ contain the domain-independent attributes.

To achieve the above goals, we introduce adversarial networks $\bm{D}_{to\mathcal{S}}$ and $\bm{D}_{to\mathcal{T}}$ for each domain-specific decoders, respectively. They attempt to determine which domain is the origin of reconstructed images. Taking $\{f_\mathcal{S}, f_\mathcal{T}, \bm{G}_{to\mathcal{T}}, \bm{D}_{to\mathcal{T}}\}$ as an example, feed $f_\mathcal{S}$ and $f_\mathcal{T}$ into $\bm{G}_{to\mathcal{T}}$, then attain $\hat i_{\mathcal{S}to\mathcal{T}}$ and $\hat i_{\mathcal{T}to\mathcal{T}}$ respectively. $\bm{D}_{to\mathcal{T}}$ aims to distinguish the domains of $\hat i_{\mathcal{S}to\mathcal{T}}$ and $\hat i_{\mathcal{T}}$. Similar to the above feature-level adversarial training, the loss of the inverse discrimination result is used to update the $\bm{G}_{c}$ and $\bm{G}_{to\mathcal{T}}$. As a result, the photo-realistic image $\hat i_{\mathcal{S}to\mathcal{T}}$ is generated to fool $\bm{D}_{to\mathcal{T}}$.

\vspace{0.15cm}
\noindent\textbf{3.1.2 \,\,Coarse Training for Crowd Counting}
\vspace{0.15cm}

After generating the translated images $\hat I_{\mathcal{S}to\mathcal{T}}$, the coarse counter $\bm{C}^{coarse}$ is trained on $\hat I_{\mathcal{S}to\mathcal{T}}$ and $A_{\mathcal{S}}$ by using the traditional supervising regression method. In practice, given a batch of translation results in each iteration of IFS, $\bm{C}^{coarse}$ will be trained once. In other word, the image translation model and the coarse counter are trained together.

\vspace{0.15cm}
\noindent\textbf{3.1.3 \,\,Loss Functions}
\vspace{0.15cm}

To train the proposed framework, in each iteration, the discriminators $\bm{D}_{c}$, $\bm{D}_{to\mathcal{S}}$ and $\bm{D}_{to\mathcal{T}}$ are updated using an adversarial loss; then update the parameters of $\bm{G}_{c}$, $\bm{G}_{to\mathcal{S}}$, $\bm{G}_{to\mathcal{T}}$, and $\bm{C}^{coarse}$ by optimizing following functions:
\begin{equation}\label{Loss:G}
\begin{aligned}
\mathcal{L}=\!\mathcal{L}^{task}\! + \alpha \mathcal{L}_{\bm{D}_{c}}^{adv}+\beta \mathcal{L}_{\bm{D}_{to\mathcal{S}}}^{adv}+\gamma \mathcal{L}_{\bm{D}_{to\mathcal{T}}}^{adv}\!+\!\mathcal{L}^{cons}\!,
\end{aligned}
\end{equation}
where the first item is task loss for counting, the middle threes are adversarial loss for three discriminators, and the last item is the consistency loss. By repeating the above training, the models will be obtained. Next, we will explain the concrete definitions of them. Note that $\theta_*$ means that the parameters of the model $*$.

\noindent\textbf{Task Loss}\quad For the counting task, we train $\bm{C}^{coarse}$ via optimizing $\mathcal{L}^{task}(\theta_{\bm{C}^{coarse}})$, a standard MSE loss.

\noindent\textbf{Feature-level Adversarial Loss}\quad To effectively extract domain-shared features, we minimize feature-level LSGAN loss \cite{mao2017least} to train $\bm{D}_{c}$. The loss and inverse loss is denoted by $\mathcal{L}_{\bm{D}_{c}}$ and $\mathcal{L}_{\bm{D}_{c}}^{adv}$, respectively.



\noindent\textbf{Multi-scale Translation Adversarial Loss}\quad We find that the traditional methods are prone to generating weird data that contain distorted color distribution. The main reason is that the adversarial training is unstable and it causes that some neurons are sensitive to specific data. To alleviate this problem, we propose a multi-scale translation adversarial loss (MS Ad for short). which adopts a full convolution discriminator to distinguishes the domains of two images under the different image scales. It is a MSE-like loss and has been proved in LSGAN \cite{mao2017least} with better stability compared with the Cross-entropy loss.
Take $\bm{D}_{toS}$ as an example, $\mathcal{L}_{\bm{D}_{to\mathcal{S}}}$ and $\mathcal{L}_{\bm{D}_{to\mathcal{S}}}^{adv}$ are formulated as:
\begin{equation}
\begin{aligned}
\mathcal{L}_{\bm{D}_{to\mathcal{S}}}(\theta_{\bm{D}_{to\mathcal{S}}})= \frac{1}{2}\sum_{l=1}^{2} \Big\{&||\bm{D}_{to\mathcal{S}}(i_{S}^{l})-0||^{2}\\ &+||\bm{D}_{toS}(i_{\mathcal{T}to\mathcal{S}}^{l})-1||^{2} \Big\},
\end{aligned}
\end{equation}
and
\begin{equation}
\begin{aligned}
\mathcal{L}_{\bm{D}_{to\mathcal{S}}}^{adv}(\theta_{\bm{G}_{c}},\theta_{\bm{G}_{to\mathcal{S}}})=\frac{1}{2}\sum_{l=1}^{2}||\bm{D}_{to\mathcal{S}}(i_{\mathcal{T}to\mathcal{S}}^{l})-0||^{2},
\end{aligned}
\end{equation}
where $l=1,2$ respectively represents the size of inputs, namely 0.5x and 1.0x. During the training, $\bm{D}_{to\mathcal{S}}$ attempts to distinguish the origins of $i_{\mathcal{S}}$ and $i_{\mathcal{T}to\mathcal{S}}$. At the same time, by optimizing $\mathcal{L}_{\bm{D}_{to\mathcal{S}}}^{adv}$, the $\bm{G}_{c}$ and $\bm{G}_{to\mathcal{S}}$ are updated to generate like-target images that can confuse $\bm{D}_{to\mathcal{S}}$. Similarly, there are $\mathcal{L}_{\bm{D}_{to\mathcal{T}}}(\theta_{\bm{D}_{to\mathcal{T}}})$ and $\mathcal{L}_{\bm{D}_{to\mathcal{T}}}^{adv}(\theta_{\bm{G}_{c}},\theta_{\bm{G}_{to\mathcal{T}}})$ to train $\bm{D}_{to\mathcal{S}}$ and $\{\bm{G}_{c}, \bm{G}_{to\mathcal{S}}\}$, respectively.

\noindent\textbf{Consistency Loss}\quad Mainstream translation methods have two data flows: recall process ($i_{\mathcal{S}}\! \rightarrow\! \hat i_{\mathcal{S}to\mathcal{S}}$) and the translation process ($i_{\mathcal{T}}\! \rightarrow\! \hat i_{\mathcal{T}to\mathcal{S}}$). For the former, the researchers \cite{zhu2017unpaired} usually regularize the data using pixel-wise consistency loss (namely L2 Loss). However, for image translation task, the ultimate goal is translating images instead of recalling images. Many works ignore the latter so that the model loses the original content and detailed features.

To remedy this problem, we attempt to design a loss function to constrain high-level image content. The L2 Loss in recall process is improper, because it only measure the pixel-wise distance, which is anti-translation operation. Therefore, we propose a content-aware consistency loss to regularize $i_{\mathcal{T}}$ and $\hat i_{\mathcal{T}to\mathcal{S}}$. To be specific, we adopt perceptual losses \cite{johnson2016perceptual} to formulate the difference of feature maps extracted by a pre-trained classification model VGG-16 \cite{simonyan2014very}, which are named as $\mathcal{L}_{\mathcal{T}to\mathcal{S}}^{cont}(\theta_{\bm{G}_{c}},\theta_{\bm{G}_{to\mathcal{S}}})$ and $\mathcal{L}_{\mathcal{S}to\mathcal{S}}^{cont}(\theta_{\bm{G}_{c}},\theta_{\bm{G}_{to\mathcal{S}}})$. It effectively maintains low-level local features and high-level crowd contents of the original image. Similarly,  there are $\mathcal{L}_{\mathcal{T}to\mathcal{T}}^{rec}(\theta_{\bm{G}_{c}},\theta_{\bm{G}_{to\mathcal{T}}})$ and $\mathcal{L}_{\mathcal{S}to\mathcal{T}}^{cont}(\theta_{\bm{G}_{c}},\theta_{\bm{G}_{to\mathcal{T}}})$ to regularize the outputs of $\bm{G}_{to\mathcal{T}}$.

Finally, $\mathcal{L}^{cons}$ in Eq. \ref{Loss:G} is the sum of the above four consistency losses.

\subsection{Gaussian-prior Reconstruction}

In the field of crowd counting, the ground-truth of density map is generated using head locations and Gaussian kernel \cite{lempitsky2010learning}. The goal of Gaussian-prior Reconstruction (GPR) is to find the most likely head locations via comparing the coarse map and the standard kernel. After this, the pseudo map is reconstructed and used to train a final counter on the target domain.

\noindent\textbf{Density Map Generation} Firstly, we briefly review the generation process of density maps in traditional supervised methods. In the field of counting, the original label form is a set of heads positions $(\bm{x},\bm{y})=\{\left(x_{1},y_{1}), \dots, (x_{N},y_{N})\right\}$. Take a sample $(x_{i},y_{i})$ as an example, it is treated as a delta function $\delta\left(x-x_{i}, y-y_{i}\right)$. Therefore, the position set can be formulated as:

\begin{equation}
H(\bm{x},\bm{y})=\sum_{i=1}^{N} \delta\left(x-x_{i}, y-y_{i}\right).
\label{delta}
\end{equation}

For getting the density map, we convolve $H(\bm{x},\bm{y})$ with a Gaussian function $G_{k,\sigma}$, where $k$ is the kernel size and $\sigma$ is the standard deviation. In practice,  $G_{k,\sigma}$ is regarded as a discrete Gaussian Window $W_{k,\sigma}$ with the size of $k \times k$. To be specific, the value of position $(u,v)$ in $W_{k,\sigma}$ is defined as $w(u,v)=e^{-D^{2}(u,v)/2\sigma^{2}}$, where $D(u,v)$ is the distance from $(u,v)$ to the window center. It is defined as:
\begin{equation}
D(u,v)\!=\!\left[((u\!-\!(k\!+\!1)\!/\!2)^{2}\!+\!(v\!-\!(k\!+\!1)\!/\!2)^{2}\right]^{1/2}.
\end{equation}
In the experiments, we set $k$ as 15 and $\sigma$ as $4$.

\begin{figure*}[t]
	\centering
	\includegraphics[width=0.85\textwidth]{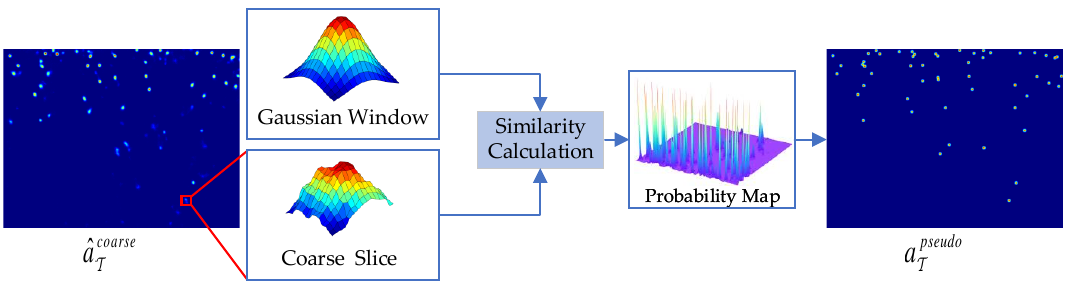}
	\caption{The generation process of pseudo labels.} \label{Fig-Gau}
\end{figure*}

\noindent\textbf{Density Map Reconstruction} A standard map is recalled according to the coarse result $\hat{a}_{\mathcal{T}}$. It consists of three steps: 1) compute probability map at the pixel level, of which each pixel represents its confidence as a Gaussian kernel's center; 2) iteratively select a maximum-probability candidate point and update the probability map in turns; 3) generate pseudo labels based on candidate points.

Here, we detailedly explain the generation of the probability map. Take a pixel $(x_i,y_i)$ in $\hat{A}_{\mathcal{T}}$ as the center, cropping a window $\hat{A}_{\mathcal{T}}^{(x_i,y_i)}$ with the size of $k \times k$. Then measuring the similarity between $\hat{A}_{\mathcal{T}}^{(x_i,y_i)}$ and $W_{k,\sigma}$ using following formulation:
\begin{equation}\label{For:P_map}
P(x_i,y_i)=\frac{1}{1+  \|\hat{A}_{\mathcal{T}}^{(x_i,y_i)}-W_{k,\sigma}\|_1},
\end{equation}
where $W_{k,\sigma}$ is a discrete Gaussian Window with the size of $k \times k$, $P(x_i,y_i) \in [0,1]$ and the higher value means that it is closer to the $W_{k,\sigma}$. Finally, the probability map $P$ is obtained. The generation flow is shown in Fig.\ref{Fig-Gau}, and the computation process is demonstrated in Algorithm \ref{al}.
\begin{algorithm}[htb]
	\caption{ Algorithm for generating pseudo labels.}
	
	\label{alg:Framwork}
	\begin{algorithmic}[1]
		\Require
		Coarse map $\hat{A}_{\mathcal{T}}$, Gaussian Window $W_{k,\sigma}$.
		\Ensure
		Pseudo label map $A_{\mathcal{T}}^{pseudo}$.
		\State
		Count the number of people, $\hat N=int(sum(\hat{A}_{\mathcal{T}}))$;
		\State
		Compute the probability map $P$ for $\hat{A}_{\mathcal{T}}^{coarse}$ with Eq. \ref{For:P_map};
		\For{$j=1$ to $\hat N$}

		\State Get a candidate point $(\hat x_{j}, \hat y_{j})=\underset{(\hat x_{j}, \hat y_{j})}{\arg \max}(P(\hat x_{j}, \hat y_{j}))$;
		\State Crop a window $\hat{A}_{\mathcal{T}}^{(\hat x_j,\hat y_j)}$ with the center $(\hat x_{j},\! \hat y_{j})$ from $\hat{A}_{\mathcal{T}}$;
		\State Update $\hat{A}_{\mathcal{T}}^{(\hat x_j,\hat y_j)} =\hat{A}_{\mathcal{T}}^{(\hat x_j,\hat y_j)} - W_{k,\sigma}$;
		
		\State Place $\hat{A}_{\mathcal{T}}^{(\hat x_j,\hat y_j)}$ back to $\hat{A}_{\mathcal{T}}$;
		\State Recompute $P$'s region where changes occur in $\hat{A}_{\mathcal{T}}$;
		\EndFor
		\State
		Generate the map $A_{\mathcal{T}}^{pseudo}$ with $\{(\hat x_{1}, \hat y_{1}), ...,(\hat x_{\hat N}, \hat y_{\hat N})\}$.\\
		\Return $A_{\mathcal{T}}^{pseudo}$.
		
	\end{algorithmic}\label{al}
\end{algorithm}

\noindent\textbf{Re-training Scheme} Although the above reconstruction can effectively prompt the density quality, it may generate a few mistaken head labels from the coarse map. In addition, its time complexity is $O(n)$, which is not efficient. To remedy these problems, we re-train a final counter $\bm{C}^{final}$ using ${I}_{\mathcal{T}}$ and ${A}_{\mathcal{T}}^{pseudo}$ based on the $\theta_{\bm{C}^{coarse}}$. The error labels will be alleviated as the model converges. During the test phase, the $\bm{C}^{final}$ is performed to directly more high-quality predictions than the coarse results.

\subsection{Network Architecture}

This section briefly describes our network architectures.  $\bm{G}_{c}$ consists of four residual blocks and outputs $512$-channel feature map with the 1/4 size of inputs. $\bm{G}_{to\mathcal{S}}$ and $\bm{G}_{to\mathcal{T}}$ have the same architecture, including six convolutional/de-convolutional layers. For the discriminators, they are all designed as a five-layer convolution network. The counters utilize the first 10 layers of VGG-16 \cite{simonyan2014very}, and up-sample to the original size via a series of de-convolutional layers. All detailed configurations of the networks are shown in supplementary materials, and the code will be released as soon as possible.

\subsection{Implementation details}

\textbf{Parameter Setting} During the training process of IFS, the weight parameters $\alpha$, $\beta$, and $\gamma$ in Eq.\ref{Loss:G} are set to $0.01$, $0.1$, and $0.1$, respectively. Due to the limited memory, in each iteration, we input $4$ source images and $4$ target images with a crop size of $480\times480$. Adam algorithm \cite{kingma2014adam} is performed to optimize the networks. The learning rate for the IFS models is set as $10^{-4}$, and the learning rate for $\bm{C}^{coarse}$ is initialized as $10^{-5}$. After $4,000$ iterations, we stop updating the IFS models, but continue to update the $\bm{C}^{coarse}$ until it converges. For GPR process, $\bm{C}^{final}$'s learning rate is set as $10^{-5}$. Our code is developed based on the $C^{3}$ Framework \cite{C3} on NVIDIA GTX $1080$Ti GPU.

\textbf{Scene Regularization} In other fields of domain adaptation, such as semantic segmentation, the object distribution in street scenes is highly consistent. Unlike this, current crowd real-world datasets are very different in terms of density. For avoiding negative adaptation, we adopt a scene regularization strategy proposed by \cite{wang2019learning}. In other word, we manually select some proper synthetic scenes from GCC as the source domain for different target domains. Due to no experiment on UCSD \cite{chen2012feature} and Mall\cite{chan2008privacy} in SE CycleGAN \cite{wang2019learning}, we define the scene regularization for them. The detailed information is shown in the supplementary.


\section{Experimental Results}
\subsection{Evaluation Criteria}

Following the convention, we utilize Mean Absolute Error (MAE) and Mean Squared Error (MSE) to measure the counting performance of models, which are defined as:
\begin{equation}
MAE = \frac{1}{N}\sum\limits_{i = 1}^N {\left| {{y_i}- {{\hat y}_i}} \right|},
MSE = \sqrt {\frac{1}{N}\sum\limits_{i = 1}^N {{{\left| {{y_i} - {{\hat y}_i}} \right|}^2}} },
\end{equation}
where $N$ is the number of images, ${{y_i}}$ is the groundtruth number of people and ${{{\hat y}_i}}$ is the estimated value for the $i$-th image. Besides, PSNR and SSIM \cite{wang2004image} is adopted to evaluate the quality of density maps.

\subsection{Datasets}
\label{dataset}
For verifying the proposed domain-adaptive method, the experiments are conducted from GCC \cite{wang2019learning} to another six real-world, namely Shanghai Tech Part A/B \cite{zhang2016single}, UCF-QNRF \cite{idrees2018composition}, WorldExpo'10 \cite{zhang2016data}, Mall \cite{chen2012feature} and UCSD \cite{chan2008privacy}.

\textbf{GCC} is a large-scale synthetic dataset, which consists of still $15,212$ images with a resolution of $1080 \times 1920$.

\textbf{Shanghai Tech Part A} is a congested crowd dataset, of which images are from a photo-sharing website. It consists of 482 images with different resolutions.

\textbf{Shanghai Tech Part B} is captured from the surveillance camera on the Nanjing Road in Shanghai, China. It contains $716$ samples with a resolution of $768 \times 1024$.

\textbf{UCF-QNRF} is an extremely congested crowd dataset, including 1,535 images collected from Internet, and annotating in 1,251,642 instances.

\textbf{WorldExpo'10} is collected from 108 surveillance cameras in Shanghai 2010 WorldExpo, which contains $3,980$ images with a size of $576 \times 720$.

\textbf{Mall} is collected using a surveillance camera installed in a shopping mall, which records the $2,000$ sequential frames with a resolution of $480 \times 640$.

\textbf{UCSD} is an outdoor single-scene dataset collected from a video camera at a pedestrian walkway, which contains $2,000$ image sequences with a size of $158 \times 238$.

\subsection{Module-level Ablation Study on Shanghai Tech Part A}
We conduct a group of detailed ablation study to verify the effectiveness of our proposed models on Shanghai Tech Part A. To be specific, the different models' configurations are explained as follows:

Table \ref{Table-ablation} reports the quantitative results of different module fusion methods.

\textbf{NoAdpt}: Train the counter on the original GCC.

\textbf{IFS-a}: Train the translated GCC of IFS w/o feature-level adversarial learning.

\textbf{IFS-b}: Train the translated GCC of IFS with feature-level adversarial learning.

\textbf{IFS-b + GPR-a}: Reconstruct pseudo labels using the results of the counter in IFS-b.

\textbf{IFS-b + GPR-b}: Retrain the counter with the pseudo labels of IFS-b + GPR-a. It is the full model of this paper, namely the proposed DACC.

\begin{table}[htbp]
	\centering
	
	\caption{The performance of the proposed different models on Shanghai Tech Part A.}
	\setlength{\tabcolsep}{1.6mm}{
		\begin{tabular}{cIc|c|c|c}
			\whline
			\multirow{2}{*}{Method}	 &\multicolumn{4}{c}{Shanghai Tech Part A} \\
			\cline{2-5}
			&  MAE &MSE &PSNR &SSIM  \\
			\whline
			NoAdpt   &206.7 &297.1 &18.64 &0.335 \\
			\hline
			IFS-a  &127.3 &190.6  &21.80 &0.458    \\
			\hline	
			IFS-b  &120.8 &184.6  &21.41 &0.466	  \\
			\hline	
			IFS-b + GPR-a &120.6 &184.4 &19.73 &\textbf{0.760} \\
			\hline
			IFS-b + GPR-b (DACC) &\textbf{112.4} &\textbf{176.9} &\textbf{21.94} &0.502  \\
			\whline
		\end{tabular}
	}\label{Table-ablation}
\end{table}

\textbf{Analysis of IFS}\,\, From the table, the methods with adaptation far exceed NoAdpt, which shows the effectiveness of domain adaptation. By comparing the results IFS-a and IFS-b, the errors are significantly reduced (MAE/MSE: from 127.3/190.6 to 120.8/184.6). It indicates that the feature-level adversarial learning effectively facilitates the segregation of inter-domain features.

\begin{figure*}[t]
	\centering
	\includegraphics[width=.98\textwidth]{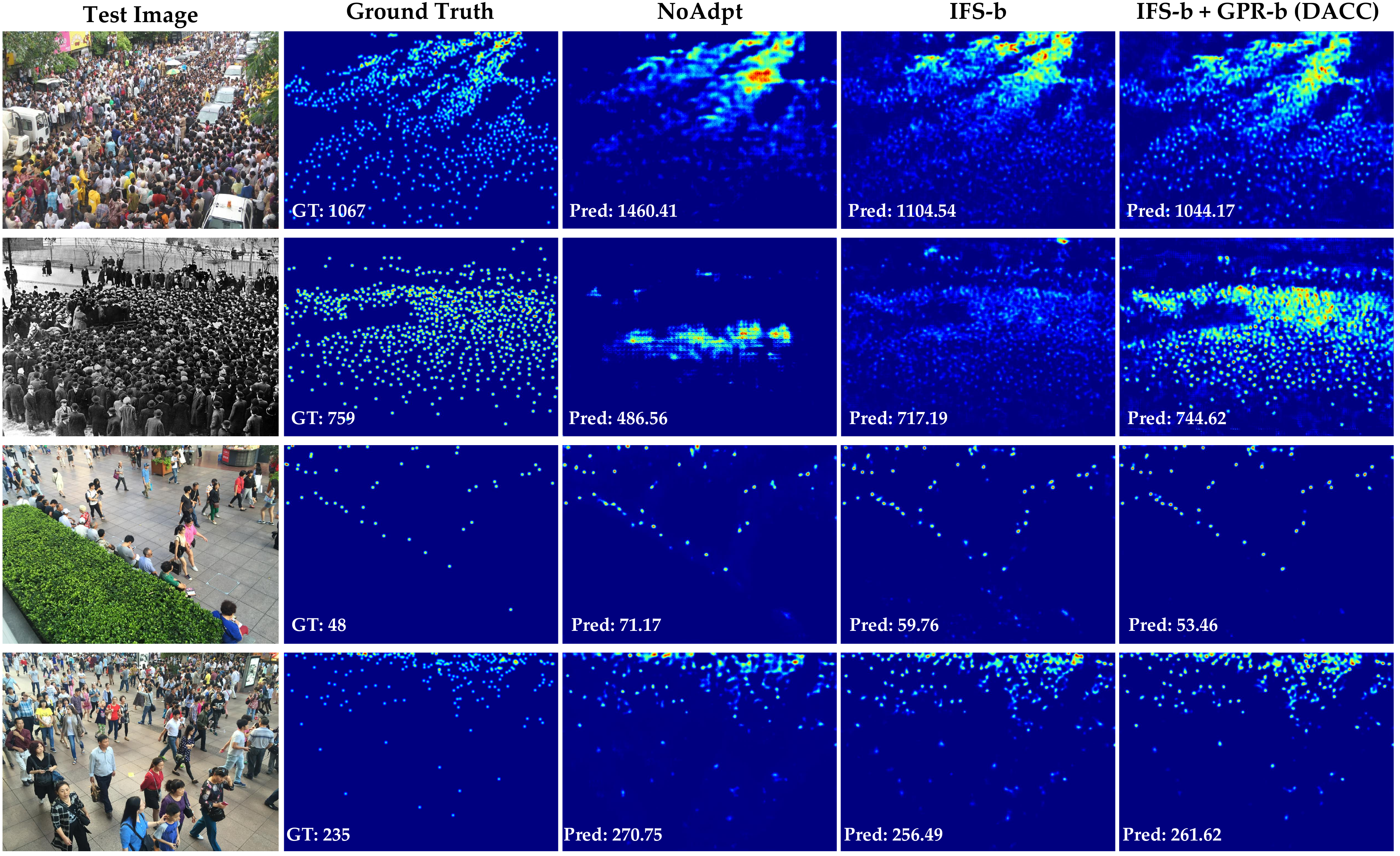}
	\caption{Exemplar results of adaptation from GCC to Shanghai Tech Part A and B dataset. In the density map. ``GT'' and ``Pred'' represent the number of ground truth and prediction, respectively. Row 1 and 2 come from ShanghaiTech Part A, and others are from Part B. }\label{re-examplars}
\end{figure*}

\begin{table*}[htbp]
	\centering
	\caption{The performance of the proposed different loss combinations on UCF-QNRF.}
	\setlength{\tabcolsep}{4mm}{
		\begin{tabular}{lIc|c|c|cIc|c|c|c}
			\whline
			\multirow{2}{*}{Loss Combinations}	 &\multicolumn{4}{cI}{CycleGAN Structure} &\multicolumn{4}{c}{IFS-b Structure}\\
			\cline{2-9}
			&  MAE &MSE &PSNR &SSIM &  MAE &MSE &PSNR &SSIM \\
			\whline
			{\color{red}Ad} + C (original)  &257.3 &400.6 &20.80 &0.480 & 243.9 &392.6 &20.77 &0.607\\
			{\color{red}MS Ad} + C & \underline{232.9} &\underline{394.0} &\underline{20.97} &\underline{0.575}&221.8  &385.9 &21.23 &0.642	  \\
			\hline
			\hline		
			ad + c + {\color{red}SE}  &230.4 &384.5 &21.03 &\textbf{\underline{0.660}}  &225.3 &281.7 &21.57 &\underline{\bfseries0.690} \\
			ad + c + {\color{red}CA}  &\underline{223.7} &\underline{381.7} &\underline{21.09} &0.612 &\underline{215.8} &\underline{361.0} &\underline{21.77} &0.676 \\
			\hline	
			\hline
			MS Ad + C + CA & \textbf{218.1} & \textbf{380.0}  &\textbf{21.17} &0.624 &\bfseries211.7 &\bfseries357.9 &\bfseries21.94 &0.687  \\
			\whline
		\end{tabular}
	}\label{Table-loss}
\end{table*}

\textbf{Analysis of GPR}\,\, When introducing GPR-a into IFS-b, the counting errors are slightly different (MAE/MSE: 120.8/184.6 v.s. 120.6/184.4). The main reason is that the rounding operation for counting number in Line 1 of Algorithm \ref{al}. It is a double-edged sword, which maybe decrease or increase the errors. The slight performance fluctuations are not important. Our concern is to improve the quality of density map and remove some misestimations by the further retraining scheme. Correspondingly, since GPR-a generates the standard pseudo labels, SSIM achieves the value of 0.760, far more than the previous result of 0.466. After retraining a final counter, the mistaken estimations in the background are dramatically suppressed. As a result, the MAE and MSE of DACC are further reduced (MAE/MSE: 120.6/184.4 v.s. 112.4/176.9).

\textbf{Visualization Results}\,\, Fig. \ref{re-examplars} shows the visualization results of the proposed step-wise models (NoAdpt, IFS-b and IFS-b + GPR-b) on Shanghai Tech Part A and B. From the results of Column 3, NoAdpt only reflects the trend of density distribution. For the second sample, NoAdpt produce a weird density map, which seems to be not consistent with the original image. The main reason is that GCC data are RGB images, but the second sample is a gray-scale scene. The NoAdpt counter fully over-fits the RGB data so that it performs poorly on gray-scale images. After introducing IFS, the visual results can show the coarse density distribution. For some sparse crowd regions (such as Row 3), the counter yields the fine density map close to the ground truth. Further, the final results of DACC present two advantages in visual perception. Firstly, DACC outputs the more precise density maps, of which points are similar to the standard Gaussian kernel. It will prompt the performance of person localization. Secondly, the mistaken estimations are effectively reduced, especially in Row 3 and 4. In general, DACC's predictions are better than those of other models' in terms of the quantitative and qualitative comparisons.

\subsection{Loss-level Ablation Study on UCF-QNRF}

\begin{table*}[htbp]
	\centering
	
	\caption{The performance of no adaptation (No Adpt), CycleGAN, SE CycleGAN, FSC, FA and the proposed methods on the six real-world datasets.}
	\setlength{\tabcolsep}{0.13cm}{
		\begin{tabular}{c|cIc|c|c|cIc|c|c|cIc|c|c|c}
			\whline
			\multirow{2}{*}{Method}	&\multirow{2}{*}{DA} &\multicolumn{4}{cI}{Shanghai Tech Part A}&\multicolumn{4}{cI}{Shanghai Tech Part B} &\multicolumn{4}{c}{UCF-QNRF}\\
			\cline{3-14}
			& & MAE &MSE &PSNR &SSIM &MAE & MSE &PSNR &SSIM &MAE & MSE &PSNR &SSIM \\
			\whline
			CycleGAN \cite{zhu2017unpaired} &\rmark &143.3 &204.3 &19.27 &0.379 &25.4 &39.7 &24.60 &0.763   &257.3 &400.6 &20.80 &0.480 \\
			\hline	
			SE CycleGAN \cite{wang2019learning}  &\rmark&123.4 &193.4 &18.61 &0.407 &19.9 &28.3 &24.78 &0.765 &230.4 &384.5 &21.03 &0.660 \\
			\hline
			SE Cycle GAN (JT) \cite{wang2020pixel} &\rmark &119.6 &189.1 &18.69 &0.429 &16.4 &25.8 &26.17 &0.786 &225.9 &385.7 &21.10 &0.642 \\
			\hline
			FSC  \cite{han2020focus}    &\rmark &129.3&  187.6 &21.58&\bfseries0.513 & 16.9 & 24.7 &26.20 &0.818  & 221.2 &390.2 & \bfseries23.10 &0.708 \\
			\hline
			FA  \cite{gao2019feature} &\rmark &144.6 &200.6 &-  &- &16.0 & 24.7 &- &-  &269.5 &407.9 &-  &-\\
            \hline
			LIDK \cite{cai2021leveraging} &\rmark &- &- &-  &- &14.3 & 22.8 &- &-  &224.3 &375.8 &-  &-\\
			\whline	
			NoAdpt (ours)  &\xmark&206.7 &297.1 &18.64 &0.335 &24.8 &34.7 &25.02 &0.722 &292.6 &450.7 &20.83 &0.565  \\
			\hline
			DACC (ours)  &\rmark  &\textbf{112.4} &\textbf{176.9} &\textbf{21.94} &0.502 &\bfseries13.1     &\bfseries19.4 &\bfseries28.03 &\bfseries0.888
			&\bfseries203.5 &\bfseries343.0 &21.99 &\bfseries0.717  \\
			\whline
		\end{tabular}
		\vspace{0.05cm}
	}
	
	\setlength{\tabcolsep}{0.10cm}{
		\begin{tabular}{c|cIc|c|c|c|c|cIc|c|c|cIc|c|c|c}
			\whline
			\multirow{2}{*}{Method}	&\multirow{2}{*}{DA}  &\multicolumn{6}{cI}{WorldExpo'10 (only MAE)} &\multicolumn{4}{c}{UCSD} &\multicolumn{4}{c}{Mall} \\
			\cline{3-16}
			&  &S1 &S2 &S3 &S4 &S5 &Avg. & MAE &MSE &PSNR &SSIM & MAE &MSE &PSNR &SSIM  \\
			\whline
			CycleGAN \cite{zhu2017unpaired} &\rmark  &4.4 &69.6 &49.9 &29.2  &9.0 &32.4 &- &- &-  &-&- &- &-  &-   \\
			\hline	
			SE CycleGAN \cite{wang2019learning} &\rmark &\bfseries4.3 &59.1 &43.7  &17.0 &7.6 &26.3 &- &- &-  &- &- &- &-  &-  \\
			\hline
			SE Cycle GAN (JT) \cite{wang2020pixel} &\rmark &\textbf{4.2} &\textbf{49.6} &\textbf{41.3}  &19.8 &\textbf{7.2} &\textbf{24.4} &- &- &-  &- &- &- &-  &- \\
			\hline
			FA  \cite{gao2019feature} &\rmark &5.7 &59.9 &19.7 &\bfseries14.5 &8.1 &21.6 &2.00 &2.43 &- &- &2.47  &3.25 &- &-   \\
			\whline
			NoAdpt (ours)  &\xmark&11.0&49.2 &72.2 &40.2&17.2 &38.0 &14.95 &15.31 &23.66 &0.909 &5.92 &6.70 &25.02 &0.886 \\
			\hline
			DACC (ours)  &\rmark&4.5 &\bfseries33.6 &\bfseries14.1 &30.4&\bfseries4.4 &\bfseries17.4 &\bfseries1.76  &\bfseries2.09 &\bfseries24.42 &\bfseries0.950 &\bfseries2.31 &\bfseries2.96 &\bfseries25.54 &\bfseries0.933
			\\
			\whline
		\end{tabular}
	}	
	
	\label{Table-DA}
\end{table*}

In Section \ref{intro}, we mentioned that our losses can significantly maintain the local patterns and texture features, especially for dense crowd. Here, we will verify our opinion by using the experiments that use different translation models (CycleGAN and our IFS-b) on the most congested dataset: UCF-QNRF. The results are reported in Table \ref{Table-loss}. Here, we describe the notations in the table: ``Ad'' means the traditional adversarial loss used by CycleGAN, and ``C'' indicates the standard consistency loss used by CycleGAN \cite{zhu2017unpaired}. SE is SSIM Embedding loss proposed by SE CycleGAN \cite{wang2019learning}. ``MS Ad'' and ``CA'' are our designed multi-scale adversarial loss and context-aware consistency loss.

By comparing the results of the two translation structures, we find the proposed IFS-b is better than CycleGAN under the same training loss combination. The former can extract more effective domain-invariant features than the latter. It evidences that two-stage translation via segregating domain-shared and domain independent features can generate more similar data to real scenes. In Section \ref{quality} and Fig. \ref{Fig-trans}, we discuss the translation quality of these the two structures.

\textbf{Ad Loss \emph{v.s.} MS Ad Loss} Different from the previous methods, we attempt to regularize translated images at two resolutions, which facilitates the generator's neurons more robust and remedy the distortion translation outputs. From  the final counting errors (MAE), the MS Ad loss has 9.5\% (Cycle GAN structure) and 9.1\% (IFS-b structure) improvements than the traditional Ad Loss.

\textbf{CA Loss \emph{v.s.} C Loss} C Loss only aims at the recall quality, which does not directly affect the translation. After introducing the CA Loss, the translation quality is prompted and the model attains a better counting performance (13.1\% and 11.5\% improvements of MAE on Cycle GAN and IFS-b structure).

\textbf{SE Loss \emph{v.s.} CA Loss} Both focus on improving the translation quality on congested regions. By the comparison of the reported results, we find CA loss is superior to SE loss in terms of counting performance. From the perspective of translation quality, SE loss has more significant effect than CA loss (0.660 \emph{v.s.} 0.612). The main reason is that SE loss mainly focuses on local structural similarity instead of high-level image contents.

\subsection{Comparison with the SOTAs on Real-world Datasets}
\setulcolor{blue}
In this section, we perform the experiments of DACC on six mainstream real-world datasets and compare the performance with other domain-adaptive counting methods, such as CycleGAN \cite{zhu2017unpaired}, SE CycleGAN \cite{wang2019learning}, FSC \cite{han2020focus}, FA \cite{gao2019feature} and LIDK \cite{cai2021leveraging}. Table \ref{Table-DA} lists the concrete four metrics (MAE$\downarrow$/MSE$\downarrow$/PSNR$\uparrow$/SSIM$\uparrow$). From it, the proposed DACC outperforms the other methods on all datasets. Take MAE as an example, DACC achieves 112.4, 13.1, 203.5, 17.4,1.76 and 2.31 on the six real-world datasets. In terms of some results on density quality, FSC is better than ours. The main reason is that FSC uses the crowd mask to align the semantic consistency. The extra label effectively reduce the estimation error in the background regions. More visualization results on other datasets are shown in the supplementary.

Compared with NoAdpt, DACC significantly reduces the counting errors. Take MAE as an example, DACC reduce the estimation errors by {\color{red}{45.7\%}}, {\color{red}{47.2\%}}, and {\color{red}{30.5\%}} on the first three dense datasets, respectively. On the sparse WorldExpo'10, UCSD and Mall datasets, DACC achieves more than 50\% improvement, {\color{red}{54.2\%}}, {\color{red}{88.2\%}}, and {\color{red}{61.0\%}} respectively.

For the results on UCSD, we find the PSNR and SSIM of density map is not good. The main reason is that the label definitions of them are different. The source domain (GCC) annotate the head position but the target domain (UCSD) annotate the center position. Nevertheless, it does not affect the evaluation for counting the number of people.

\section{Discussions}

\subsection{Analysis of Translated Image Quality with SOTAs}
\label{quality}
\begin{figure}
	\centering
	\includegraphics[width=0.48\textwidth]{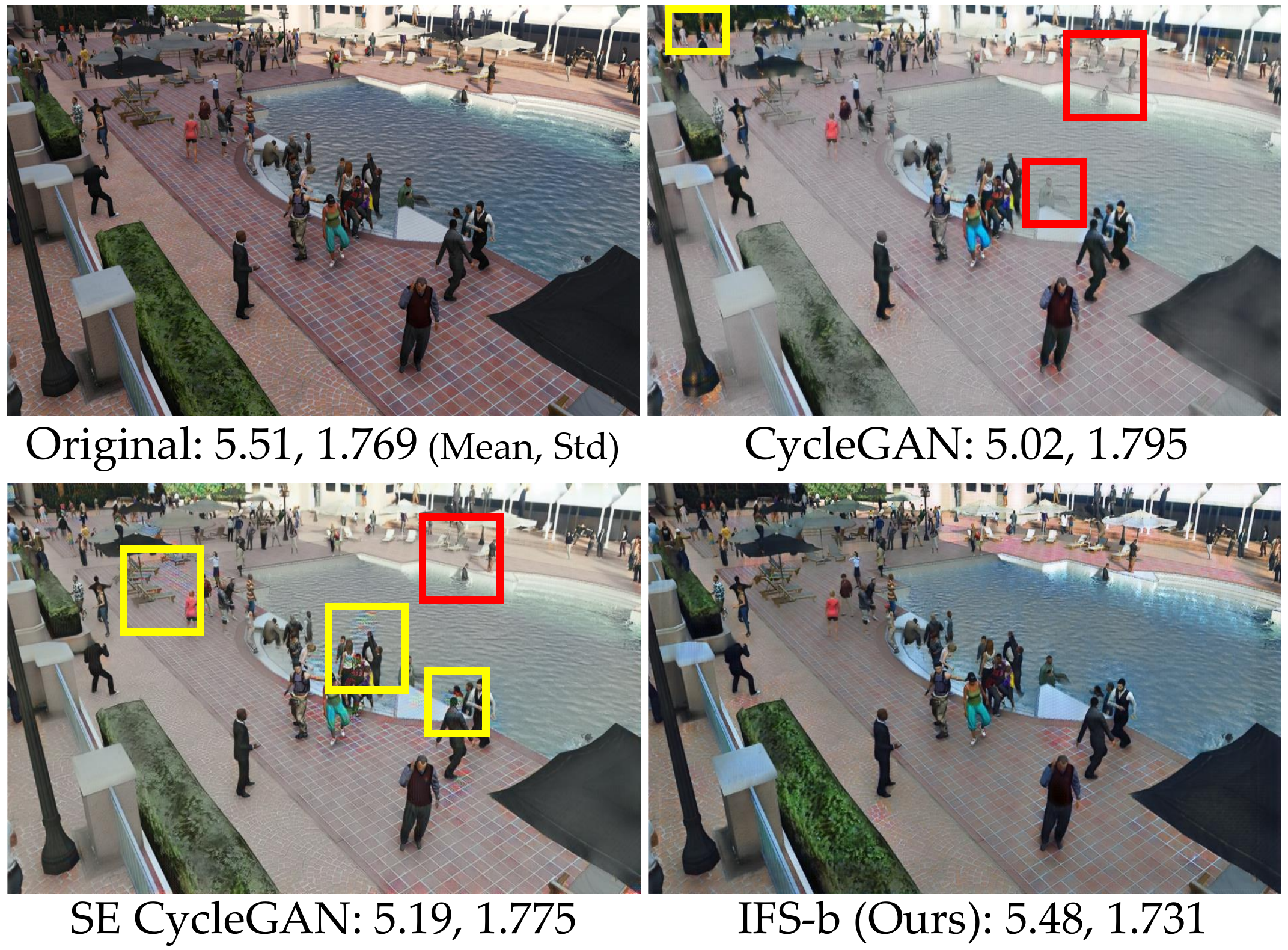}
	\caption{Comparisons of the adaptation on GCC$\rightarrow$ShanghaiTech Part B. }\label{Fig-trans}
\end{figure}

This section compares the translation results by visualization and image quality. Fig. \ref{Fig-trans} demonstrates the three results of CycleGAN, SE CycleGAN and our IFS-b. For the first two methods, they lose the key content and some detailed information, especially in the region red boxes. In addition, they also yield some distorted region in yellow boxes. In general, IFS-b maintains the crowd content well.

\begin{table}[htbp]
	\centering
	
	\caption{The quality comparison of the translated images.}
			\setlength{\tabcolsep}{0.7cm}{\begin{tabular}{c|c|c}
			\whline
			Methods &Mean $\uparrow$ &Std $\downarrow$		\\	
			\whline
			\rowcolor{mygray} NoAdpt  &5.467 &1.757 \\
			\whline
			CycleGAN   &4.935 &1.848 \\
			\hline
			SE CycleGAN &5.041 &1.846 \\
			\whline	
			DACC(ours) &\textbf{5.244} &\textbf{1.810} \\
			\whline
		\end{tabular}
  }
    \label{Table-nima}
	\end{table}

\begin{table}[htbp]
	\centering
	
	\caption{Performance of the exchange experiments (MAE/MSE).}
		\vspace{3mm}
		\setlength{\tabcolsep}{3mm}{
			\begin{tabular}{cIc}
				\whline
				Data flow: GCC$\rightarrow$Mall&Data flow: UCSD$\rightarrow$Mall \\
				\whline
				EXP1:\, 2.31/2.96&EXP2:\, 2.85/3.55 \\
				\hline
				EXP1': 2.21/2.85 &EXP2': 2.97/3.76 \\
				\whline	
		\end{tabular}}\label{tb-exchange}
\end{table}

As we all know, evaluating the translation closeness to the target domain is difficult because there is no reference image. Thus, we only assess the translation data from the perspective of image quality. Specifically, we utilize a Neural Image Assessment (NIMA) \cite{talebi2018nima}, which rates images with a mean score and a standard deviation (``std'' for short). Table \ref{Table-nima} reports these two metrics of CycleGAN, SE CycleGAN and the proposed IFS-b on GCC$\rightarrow$ShanghaiTech B. We find DACC is better than other translation methods. We also show the NoAdpt results, of which images are the original synthetic GCC data. From the scores of single image in Fig. \ref{Fig-trans}, IFS-b also outperforms CycleGAN-style methods.

\subsection{Visual Analysis of Task Performance with SOTA}
\begin{figure*}[t]
	\centering
	\includegraphics[width=.98\textwidth]{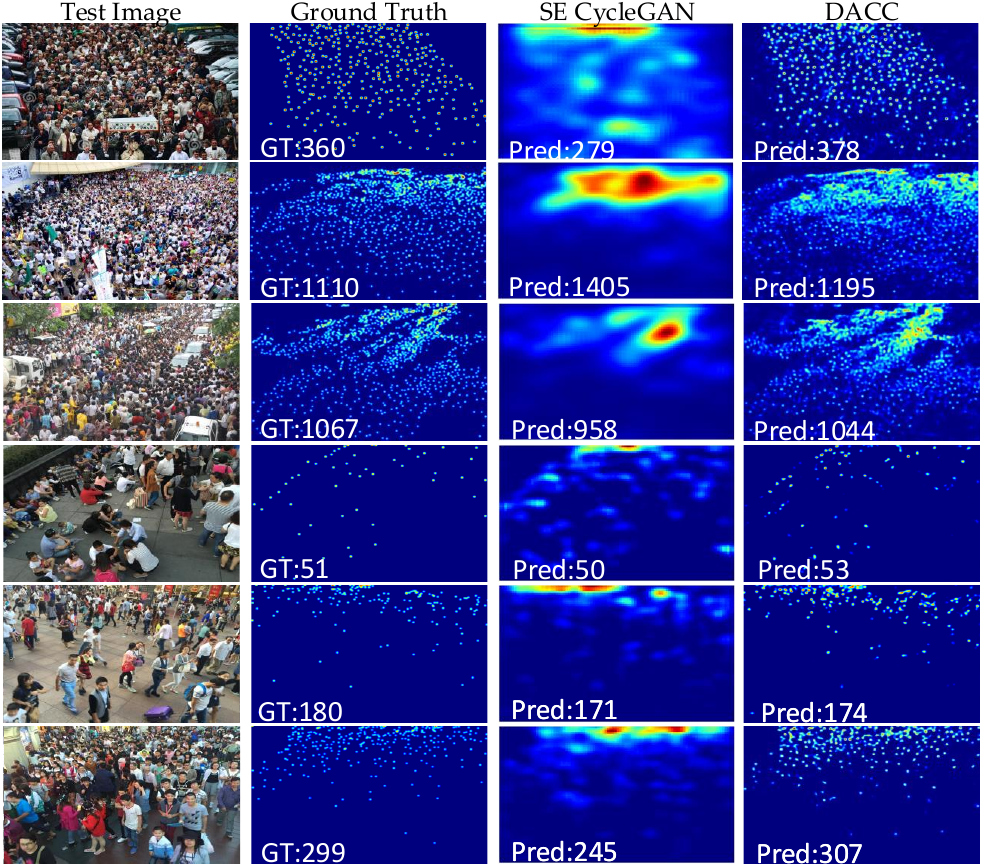}
	\caption{Exemplar results of adaptation from GCC to Shanghai Tech Part A and B dataset. In the density map. ``GT'' and ``Pred'' represent the number of ground truth and prediction, respectively. Row 1 and 2 come from ShanghaiTech Part A, and others are from Part B. }\label{com-examplars}
\end{figure*}

To vividly show the effectiveness of DACC, we compare the visual results with the SE CycleGAN \cite{wang2019learning}. It is emphasized that SE CycleGAN \cite{wang2019learning} provides visualizations of its counting results on Shanghai Tech Part A and Shanghai Tech Part B. Soo it can be gained directly and compared with the proposed DACC's counting results. Fig. \ref{com-examplars} shows the task performance on Shanghai Tech Part A dataset with two methods, which shows the density maps predicted by DACC are highly similar to the ground truth while SE CycleGAN outputs very coarse density maps. This comparison illustrates that the proposed DACC is better than SE CycleGAN in both quality and accuracy.

\subsection{Effectiveness of IFS}

In Section \ref{IFS}, it is mentioned that IFS can effectively separate domain-shared and domain-independent features. Here, we evidence this thought by two groups of exchange experiments. To be specific, select two adaptations with the same target domain, then fix the data and exchange IFS models to translate images. Take two experiments as the examples, 1)EXP1: GCC$\rightarrow$Mall and 2)EXP2: UCSD$\rightarrow$Mall. We hope to translate GCC data in EXP1 to like-Mall images using the IFS models of EXP2. Then getting the final counter by the translated images and GPR. Finally, the evaluation is conducted on the target data, namely Mall. The above experiment is defined as EXP1'. And vice versa, the other exchange way is named as EXP2'.

\begin{figure}
	\centering
	\includegraphics[width=0.48\textwidth]{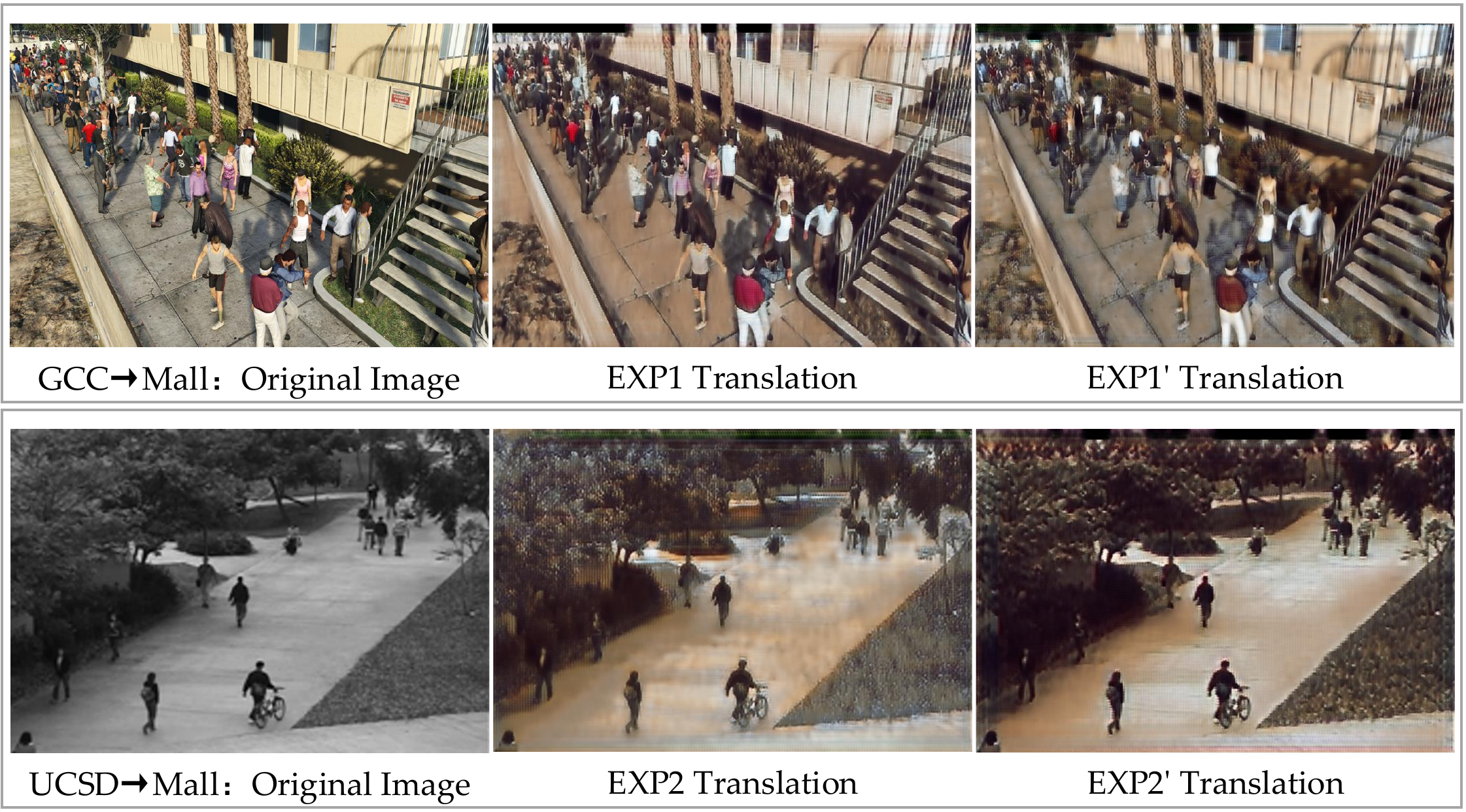}
	\caption{Visual comparison of the model exchange experiments. }\label{Fig-ifs}
\end{figure}

The counting results are listed in Table \ref{tb-exchange} and, the translation exemplars are shown in Fig. \ref{Fig-ifs}. From them, we find that: given source and target data, exchanging IFS models barely affects the performance of crowd counting and image translation.

\subsection{The performance of counter $\pmb{C}$  via Supervised Learning}
In our work, the core is not to design a crowd counter, so we do not pay much attention to the supervised performance in the target domain. However, in order to prove that the IFS image translation proposed in this paper can effectively reduce the domain gap, we conduct supervised training on several target domains. Table \ref{Table-data} compares the performance of counter $\bm{C}$ between the supervised training in the target domain and domain adaptation. As shown in Table \ref{Table-data}, the MAE and MSE of the counter used in this paper are 69.6 and 125.9 respectively on Shanghai Tech Part A, and the MAE and MSE of supervised training on Shanghai Tech Part B are 8.1 and 14.1, respectively. The results in the table show that there is a large gap between no domain adaptation and supervised training, which is significantly reduced after domain adaptation.

\begin{table}[htbp]
	
	\centering	
	\caption{The comparison of $\bm{C}$ in the proposed adaptation method and supervised training.}
	\setlength{\tabcolsep}{1.6mm}{
		\begin{tabular}{c|c|c|cc|cc}
			\whline
			\multirow{2}{*}{Methods}&\multirow{2}{*}{DA}&\multirow{2}{*}{T-GT} &\multicolumn{2}{c|}{SHT A} &   \multicolumn{2}{c}{SHT B}\\
			\cline{4-7}
			&&&               MAE    &MSE   &MAE &MSE		\\
			\whline
			NoAdapt    &\xmark&\xmark&206.7 &297.1 &24.8&34.7\\
			\hline
			DACC(ours) &\rmark&\xmark&\underline{112.4} &\underline{176.9} &\underline{13.1} & \underline{19.4}\\
			\whline
			\rowcolor{mygray}Supervised&\xmark&\rmark &\textbf{69.6} &\textbf{125.9}&\textbf{8.1}&\textbf{14.1}\\
			\whline	
		\end{tabular}
	}\label{Table-data}
\end{table}

\begin{figure*}[t]
	\centering
	\includegraphics[width=0.98\textwidth]{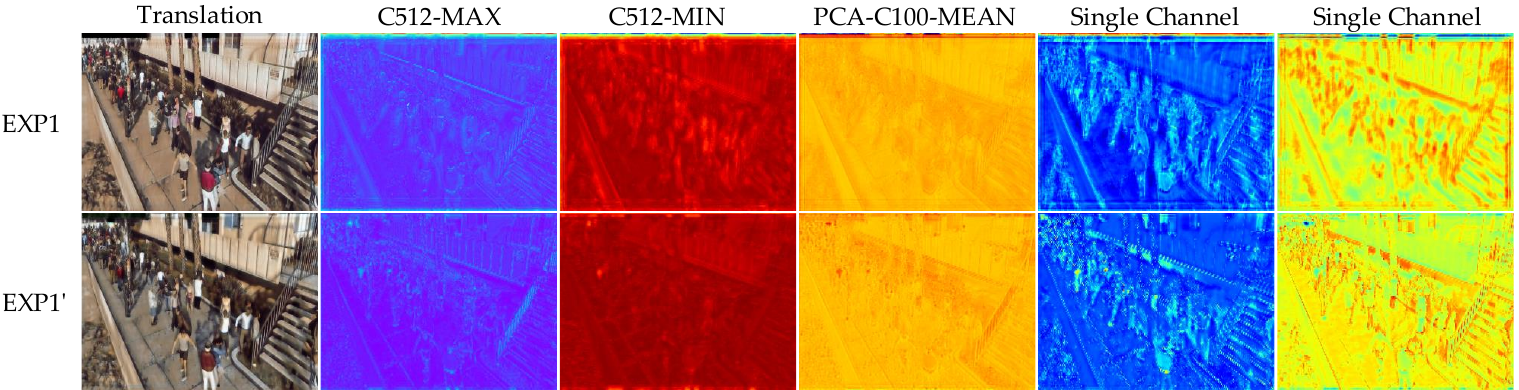}
	\caption{The feature visualization of $\bm{G}_c$ in EXP1 and EXP1' with the same source image.} \label{Fig-feature}
\end{figure*}
\subsection{IFS-b Domain-shared Feature Visualization }

In order to verify the effectiveness of our proposed IFS, we conduct a group of exchange experiments in Section 4.5. Here, we show the visualization results at the feature level. To be specific, the domain-shared features of $\bm{G}_c$ in EXP1 and EXP1' are illustrated in Fig. \ref{Fig-feature}. The first column denotes the image translation results, and the second and third columns respectively represent the maximum, minimum values of each pixel in 512 channels. The fourth column is the average value of each pixel after reducing the original features to 100 channels via PCA. The last two are some similar features selected from 512-channel feature maps. From these visualization results, we find that different $\bm{G}_c$ from EXP1 and EXP1' can extract similar features for the same image. From Column 5 and 6, there are high responses for the crowd region. In a word, these results evidence that the proposed IFS can extract domain-shared crowd contents.

\section{Conclusion}
In this paper, we present a Domain-Adaptive Crowd Counting (DACC) approach without any manual label. Firstly, DACC translates synthetic data to high-quality photo-realistic images by the proposed Inter-domain Features Segregation (IFS). At the same time, we train a coarse counter on translated images. Then, Gaussian-prior Reconstruction (GPR) generate the pseudo labels according to the coarse results. By the re-training scheme, a final counter is obtained, which further refines the quality of density maps on real data. Experimental results demonstrate that the proposed DACC outperforms other state-of-the-art methods for the same task. In future work, we plan to extend IFS on multiple domains so that it can extract more effective and robust crowd contents to improve the counting performance.


\bibliographystyle{IEEEtran}
\bibliography{IEEEabrv,reference}

\end{document}